\documentclass[runningheads]{llncs}

\usepackage{graphicx}
\usepackage{amsmath,amssymb}
\usepackage{hyperref}
\usepackage{float}

\begin{document}

\title{A Multi-Axial Mindset for Ontology Design: Lessons from Wikidata’s Polyhierarchical Structure}
\titlerunning{Multi-Axial Ontology Design in Wikidata}

\author{Ege Atacan Do\u{g}an, Peter F. Patel-Schneider}
\authorrunning{E. A. Do\u{g}an, Peter F. Patel-Schneider}
\institute{Julius-Maximilians-Universität Würzburg, \email{egeatacandogan@gmail.com} \\ Independent Researcher, \email{pfpschneider@gmail.com}}

\maketitle

\begin{abstract}
Traditional ontology design emphasizes disjoint and exhaustive top-level distinctions such as continuant vs.\ occurrent, abstract vs.\ concrete, or type vs.\ instance. These distinctions are used to structure unified hierarchies where every entity is classified under a single upper-level category. Wikidata, by contrast, does not enforce a singular foundational taxonomy. Instead, it accommodates multiple classification axes simultaneously under the shared root class \texttt{entity}. This paper analyzes the structural implications of Wikidata’s polyhierarchical and multi-axial design. The Wikidata architecture enables a scalable and modular approach to ontology construction, especially suited to collaborative and evolving knowledge graphs.
\end{abstract}

\footnotetext{This paper was prepared with assistance from ChatGPT-4. The model supported drafting, structuring, and refining arguments. All interpretations and conclusions remain the responsibility of the authors.}

\section{Introduction}

Ontology design has traditionally adhered to disjoint and exhaustive top-level distinctions, such as \emph{continuant}/\emph{occurrent}, or \emph{abstract}/\emph{concrete}. These distinctions are what we call the \texttt{primary split} of the ontology.  Foundational ontologies like BFO, DOLCE, and SUMO then extend this top-level split to a tree-shaped upper ontology grounded mostly in binary oppositions. 

Wikidata, by contrast, adopts a structurally different approach. Rather than committing to a single primary split, it organizes knowledge using multiple classification axes simultaneously. The root class \texttt{entity} (Q35120) serves as the starting point for several conceptually distinct and sometimes overlapping trees, such as the \emph{abstract}/\emph{concrete} axis, the \emph{individual}/\emph{collective} axis, and the \emph{observable}/\emph{unobservable} axis. These axes consist of mutually disjoint and complete (exhaustive) classes. However, a given class in one axis is not disjoint with a class in another axis. While having many direct subclasses under \texttt{entity} may seem odd compared to the sparseness of other foundational ontologies, this can work well in Wikidata’s flexible structure.

This paper presents a structural analysis of Wikidata’s ontology and introduces the concept of a \textit{multi-axial mindset} for ontology design. We argue that the core difference lies in the architectural design: Wikidata accommodates multiple orthogonal classification axes, permitting non-exclusive categorizations at all levels of the hierarchy. This allows entities and classes to participate in several high-level taxonomies simultaneously, in contrast to foundational ontologies that impose a single primary split.

\section{Top-Level Ontological Distinctions}
A primary split in an ontology is the initial, high-level division that structures the class hierarchy in the ontology. It defines the most fundamental categories under which all other entities are subsumed. These splits are typically exhaustive and disjoint: every entity is expected to fall under exactly one branch, and no entity should belong to more than one. The choice of primary split reflects philosophical assumptions about the nature of reality, such as whether time, materiality, or identity is foundational. Logically, any distinction within an ontology can be a primary split, even a trivial one such as \emph{apples}/\emph{non-apples}, although the intuitive understanding of classification makes this approach non-sensical. Therefore, all primary splits should have a philosophical basis.

Foundational ontologies such as BFO \cite{arp2015building}, DOLCE \cite{masolo2003dolce}, SUMO \cite{niles2001ontology}, UFO \cite{guizzardi2005ontological}\footnote{UFO has 3 other axes than given in the table, but Type/Individual is the one handled at the first split under ``thing''.}, and Cyc \cite{kishore2004computational} structure their class hierarchies according to disjoint and exhaustive top-level distinctions. These distinctions are meant to be the primary split of said ontology. Wikidata, by contrast, supports a flexible, multi-axial design in which such distinctions coexist as overlapping, non-exclusive axes under a shared root class: \texttt{entity} (Q35120).

Table~\ref{tab:foundational-ontologies} compares key ontologies along their primary splits.

\begin{table}[h]
\centering
\caption{Comparison of Top-Level Ontological Distinctions}
\label{tab:foundational-ontologies}
\begin{tabular}{|l|p{9cm}|p{9cm}|p{12cm}|}
\hline
\textbf{Ontology} & \textbf{Primary Split(s)} \\
\hline
BFO  & Continuant / Occurrent \\
\hline
DOLCE & Endurant / Perdurant / Quality / Abstract \\
\hline
SUMO & Physical / Abstract\\
\hline
UFO & Type / Individual\\
\hline
Cyc & Individual Object / Intangible / Represented Thing\\
\hline
\end{tabular}
\end{table}

These ontologies are based on a classification such that every entity belongs to exactly one subclass at almost all splits. Classification thus generally proceeds through mutually exclusive branches, resulting in a sparse upper ontology with little or no overlap between the classes there.

Wikidata's approach departs fundamentally from this pattern. It supports classification along multiple orthogonal axes, such as \emph{abstract}/\emph{concrete}, \emph{named}/\-\emph{unnamed}, and \emph{observable}/\emph{unobservable}. Instead of encoding such distinctions within a single-axis taxonomy, they are implemented as parallel subclass trees rooted in \texttt{entity}. For example, the entity \texttt{human} (Q5) can simultaneously be a \texttt{concrete object}, an \texttt{individual entity}, and an \texttt{observable entity}, without contradiction.

By not having a single disjoint and exhaustive primary split, but instead having multiple independent primary axes, Wikidata functions as a modular ontology architecture. New classification axes can be added without refactoring the entire ontology. This also means that, if a trivial split such as \emph{apples}/\emph{non-apples} is added, it does not create any problems. This approach stands in contrast to systems like BFO, which encode orthogonal meta-properties like dimensionality through repeated subclassing, leading to duplication in modeling \cite{arp2015building} where dimensionality is evaluated further down in the tree and therefore classification of dimensions is made several times. A multi-axial approach would have handled that distinction as a separate primary axis without the need for duplication.

\section{Philosophical Context}
Much of formal ontology design is shaped not only by technical requirements but also by deep philosophical commitments. Foundational ontologies generally base their primary split on a philosophically relevant axis that is deemed as somehow central. Wikidata, by contrast, is structurally pluralist. It does not privilege any single top-level distinction, such as \emph{continuant}/\emph{occurrent} or \emph{abstract}/\emph{concrete}. Instead, it supports multiple coexisting classification axes, applied partially and contextually across the graph. This design, by way of including many different conceptual facts about an object, is therefore an implementation of multiplicative ontology, as opposed to a reductionist ontology.\cite{Masolo2003WonderWeb}

\subsection*{Comparative Ontology and Framework Pluralism}

Wikidata's structure is not an instance of ontological relativism, where all conceptualizations are treated as equally valid. Rather, Wikidata exhibits what might be called \emph{framework pluralism}. In the spirit of Carnap's internal-external distinction \cite{carnap1950}, multiple ontological frameworks coexist internally in Wikidata, each guided by its own organizing principles. These frameworks are not collapsed into a single hierarchy but instead remain independent. 
There can be many reasons that a distinction under entity is considered a primary axis.  As Wikidata's structure can accommodate multiple primary axes, Wikidata therefore dodges the question of what the most important among these axes should be.

\subsection*{Epistemic pluralism on Wikidata}
Importantly, the pluralism stemming from multi-axial typing is \emph{structural}, not \emph{epistemic}. Wikidata can also support epistemic pluralism\cite{vrandecic2014wikidata} through the use of qualifiers. For example, the item for \texttt{Earth} (Q2) contains a scientifically supported inception statement (4540 million years BCE), but this has a qualifier stating that this statement is disputed by Young Earth creationism. While Wikidata can contain many different worldviews at once, it does not treat all viewpoints or theories as equally justified. Statements that are based in significant scientific and academic consensus will be marked as preferred, and will be used in contradictory cases. This approach partially echoes John Sowa’s\cite{sowa2000knowledge}, where different domains adopt distinct ontologies that can be compared but not fully unified. This section clarifies the distinction between structural and epistemic pluralism to make clear that multi-axial typing is compatible with both an epistemic monist and an epistemic pluralist approach of modeling. 

\subsection*{Reconsidering Universals and Classes}
In foundational systems like BFO, classes are typically constrained to represent universals: entities that exist across multiple instances and reflect real patterns in the world. A class, in this framework, is only legitimate if it corresponds to something metaphysically repeatable and causally unified. While such constraints enforce coherence, they also limit ontological flexibility and create a barrier between a set-based understanding of classes in which all items fitting a certain significant or insignificant pattern can be represented.

Wikidata adopts a broader understanding of what constitutes a class. Formally, classes in Wikidata resemble sets, with potentially fuzzy, partial, or context-dependent membership. Classes do not require metaphysical grounding in universals. Instead, Wikidata allows any concept to be treated as a class, provided the class serves a useful modeling function. This includes unconventional or context-sensitive categories like \emph{former fan of ABBA}\cite{arp2015building}, which does not make sense when treated as the implementation of a universal, but makes perfect sense when treated as a set-based class, instantiating all elements of the set \emph{former fans of ABBA}. Such an approach benefits from the intersection of multiple classification axes. In doing so, Wikidata better reflects the structure of human knowledge as it is used and understood across different domains.

\subsection*{Inferred Classes}

Formal ontology modeling typically privileges essential features in defining classes. A red pencil, for example, is typed as an instance of \emph{pencil}, with its color given via a property (\texttt{P462} = red), rather than being classified as a \emph{red object}. Traits like color, temperature, or shape are thus modeled as attributes, not taxonomic categories. Yet in many contexts such as art, pedagogy, and sensory analysis, feature-based classification is meaningful. Groupings such as \emph{red objects}, \emph{hot objects}, or \emph{round objects} are common in natural language. 

While these groupings are unconventional in an ontology, Wikidata’s multi-axial design permits such axes to coexist. Feature values modeled as properties can be inferred to create instance relationships to classes. This would create an explosion of statements if done on Wikidata, so it is better practice to do this at necessity when doing logical inferencing. 

A red pencil can thus be both a \emph{pencil} and an instance of \emph{red object}. While \emph{pencil} reflects a specific functional class, \emph{red object} defines a different type of grouping, directly under concrete object. In a color-based axis, a red pencil is grouped with a red car but not with a green pencil. This may not make sense in usual ontologies, but within the axis of color, it is the useful perspective.

Wikidata permits different stances on modeling in different parts of its ontology.  For example, all humans on Wikidata are supposed to be instances of \texttt{human} (Q5) and no subclass of it, so this part of Wikidata follows the universals design methodology. Other features, such as gender, occupation, and title, are represented only with properties. 

Wikidata has a large number of subclasses of human, including classes like \emph{man} and \emph{plumber}.  If Wikidata had a means of providing machine-readable definitions of these classes and inferring their instances, such as inferring that an instance of human with occupation plumber is an instance of plumber, this island of universals would easily fit within the rest of Wikidata.  Unfortunately, this facility does not exist in Wikidata, leaving it to users of Wikidata to understand how this island is set up.

\section{Multi-Axiality Beyond the Root}

While Wikidata’s class root, \texttt{entity} (Q35120), hosts many of the most prominent classification axes, the multi-axial design pattern is not confined to the top level. In practice, parallel and sometimes orthogonal axes are used throughout the ontology, even deep within domain-specific hierarchies.

\subsection*{Object}

The class \texttt{object} (Q488383), a high-level subclass of \texttt{entity}, exhibits multi-axial structure through at least two distinct disjoint partitions The \emph{natural}/\emph{artificial} axis, and the symmetry axis, \emph{symmetrical}/\emph{asymmetrical}.

\subsection*{Vehicle}

A more domain-focused example is found under the class \texttt{vehicle}. The orthogonal axes are the disjoint union between \emph{manned}/\emph{unmanned} vehicles, and the non-disjoint union between \emph{land vehicle}/ \emph{watercraft}/\emph{aircraft}/\emph{spacecraft}.

\subsection*{Dictionary}
Dictionary has 2 axes. It is union of \emph{descriptive dictionary} and \emph{prescriptive dictionary} and disjoint union of \emph{general dictionary} and \emph{specialized dictionary}.

\section{Structural Capacity versus Current Practice}

While Wikidata’s ontology accommodates multi-axial modeling in principle, its actual usage across the graph is inconsistent and often incomplete. The presence of disjoint classification axes, both at the root and throughout the ontology, demonstrates that the system is structurally capable of supporting orthogonal classification. However, this capacity is not yet realized in a systematic or coherent way.

\subsection*{Incomplete or Incoherent Axis Application}

Many conceptual axes defined under \texttt{entity} or other high-level classes are sparsely applied. In some cases, only a small proportion of relevant entities are typed along an axis.

\subsection*{Current multi-axial classes}
To evaluate multi-axial modeling in Wikidata, we queried items with more than one union statement, \texttt{P2738} (disjoint union of) or \texttt{P2737} (union of), each qualified with \texttt{list of values as qualifiers} (Q23766486). We didn't disregard deprecated statements, since the controversial axes of some items, such as some of the axes of \emph{entity}, have been deprecated.\footnote{Wikidata statements can be marked as \emph{preferred} or \emph{deprecated}, or can stay unmarked. Deprecated statements are excluded from truthy dumps and queries unless explicitly stated. We didn't use the statements under entity directly, so we used the truthy dump.}

The SPARQL query used was:

\begin{verbatim}
SELECT ?item (COUNT(*) AS ?totalUnions) WHERE {
  {?item p:P2738 ?stmt1.
   ?stmt1 ps:P2738 wd:Q23766486.}
  UNION
  {?item p:P2737 ?stmt2.
   ?stmt2 ps:P2737 wd:Q23766486.} }
GROUP BY ?item HAVING (?totalUnions > 1) ORDER BY DESC(?totalUnions)
\end{verbatim}

This query yielded a total of \emph{105 items}. Manually inspecting all these classes would be a worthwhile effort for the community. 

Some interesting discussions can be derived from the results:

\subsection*{Triangle}

The class \emph{triangle} is categorized along five conceptual axes in Wikidata, though three of these are redundant reformulations of the other two. The two axes are, \emph{equilateral}/\emph{non-equilateral} triangles, and the classification into \emph{right}/\emph{acute}/\emph{obtuse} triangles. These axes are not orthogonal, since right and obtuse triangles cannot be equilateral, the equilateral/non-equilateral distinction becomes relevant only within acute triangles.

Despite this lack of orthogonality, both distinctions are commonly used in natural language to describe triangles broadly. In everyday language and educational settings, triangles are frequently categorized as either equilateral or non-equilateral, without restricting this distinction to acute triangles. Thus, while the axes are formally entangled, their promotion to primary splits under triangle may still be justifiable on the basis of linguistic relevance.

\subsection*{Animal}

The class \emph{animal} is divided along three axes in Wikidata. Two of these correspond to taxonomic distinctions grounded in formal biological classification (Porifera and Eumetazoa, Vertebrata and invertebrate). The third axis, however, separates \emph{human animals} from \emph{non-human animals}.

From a logical or ontological standpoint, this resembles the arbitrary and structurally unprincipled split of \emph{apples}/\emph{non-apples}. However, the human/non-human distinction is widely deployed in natural language, legal frameworks, ethical discourse, and sociology. It would be equally possible to divide animals into \emph{whales}/\emph{non-whales} from a technical standpoint, but this split wouldn't have significance in any of these fields. While outside the scope of this paper, it would be worthy to investigate if the primacy of this axis is driven by anthropocentrism.

\section{Case Studies: Multi-Axial Typing in Practice}

To illustrate the structural dynamics of multi-axial classification in Wikidata, we examine in detail the following classes: \texttt{human} (Q5) and \texttt{painting} (Q3305213). These examples highlight how classification axes can be combined or omitted in different ways, and how instance-level semantics differ from class-level generalization.

\subsection*{Q5: Human}

The class \texttt{human} (Q5) represents the biological class of human beings and is used as the class for all individual human items (e.g., Q937 for Albert Einstein).

This class is classified using the following axes:
\begin{itemize}
    \item \textbf{Abstract / Concrete:} \texttt{human} is a subclass of \texttt{concrete object} (Q8205328)
    \item \textbf{Individual / Collective:} Typed as an \texttt{individual entity} (Q28160321)
    \item \textbf{Observable / Unobservable:} Human is currently not typed as an \texttt{observable entity} (Q3249551), but ideally, it should.
\end{itemize}

\noindent
These classifications are non-contradictory and stem from orthogonal axes. They reflect an entity that is physically instantiated, epistemically accessible, and numerically countable as individuals.

The case of \texttt{human} also demonstrates partial axis completeness at the class level. Since most individual humans are named, Q5 can be subclass of \texttt{named entity}, and any unnamed human will be given as an explicit exception to that classification. However, with a classification axis such as \emph{current entity}/\emph{former entity}, it does not make sense to model human in either. If we consider dead humans as former entities, and living humans as current entities, the split of temporality hasn't been made at the level of Q5 yet. ``Living human'' (as a subclass of human) would be more complete than ``human'' in terms of primary axes.

The different possible visualisations that take all these axes into account are as follows:
\begin{figure}[H]
    \centering
    \includegraphics[width=1\linewidth]{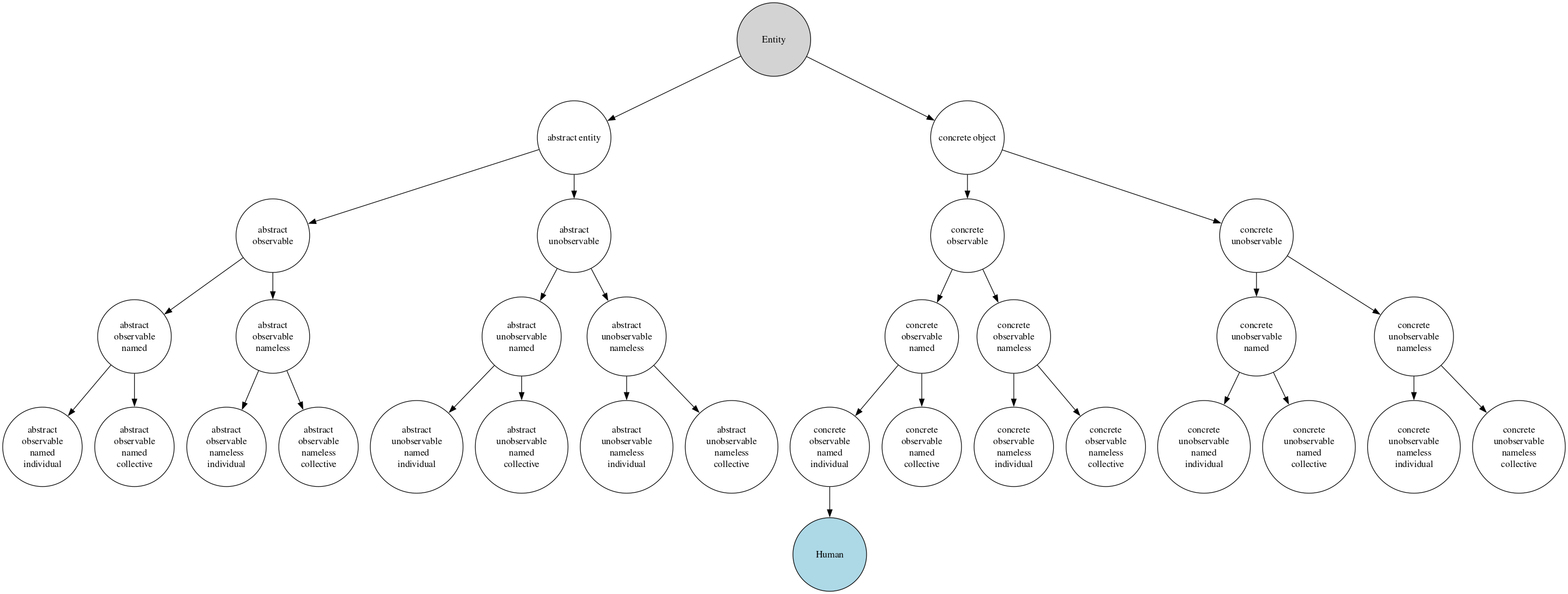}
    \caption{All 4 splits are implemented separately, leading to exponential growth.}
    \label{fig:enter-label}
\end{figure}

\begin{figure}[H]
    \centering
    \includegraphics[width=1\linewidth]{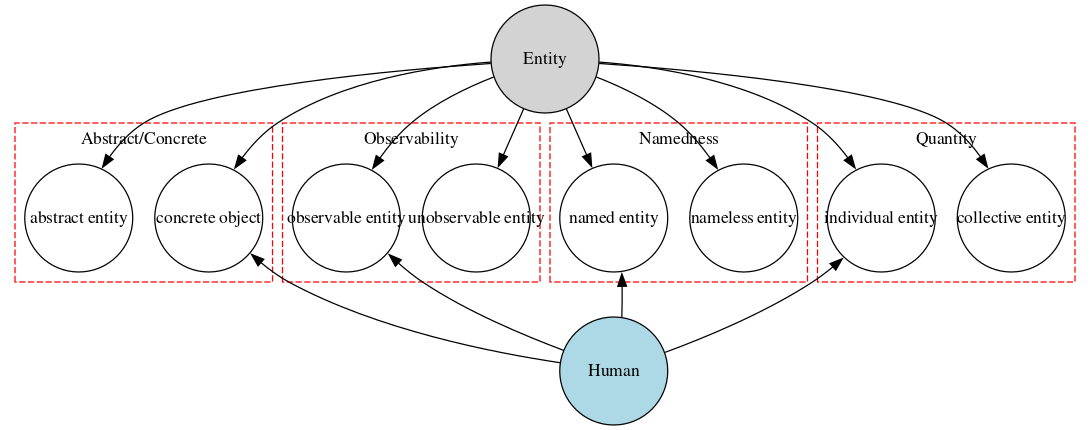}
    \caption{The 4 splits are all treated as primary splits under entity.}
    \label{fig:enter-label2}
\end{figure}

\subsection*{Q3305213: Painting}

The entity \texttt{painting} (Q3305213) represents the class of visual art works created using paint on a surface. Unlike \texttt{human}, its classification is more ambiguous across some axes:

\begin{itemize}
    \item \textbf{Abstract / Concrete:} Not classified under either.
    \item \textbf{Named / Nameless:} Not classified under either.
    \item \textbf{Observable / Unobservable:} Not explicitly typed as either, but technically should be classified under observable.
    \item \textbf{Temporal:} Not classified under the temporal axis.
\end{itemize}

This example shows how partial classification is tolerated. While many individual paintings are named (e.g., “Mona Lisa”) and current (ie. have not been destroyed), the class itself generalizes across those features. A painting may be anonymous, lost, or known only by description, making explicit classification difficult or overly restrictive at the class level.

Thus, Q3305213 illustrates the principle of allowing instance-level exhaustiveness without enforcing subclass duplication at the class level. It also underscores the multi-axial mindset: classification should follow accuracy and applicability, not abstract completeness.

\subsection*{Comparison and Implications}

The analysis of Q5 and Q3305213 thus demonstrates how Wikidata’s ontology supports full multi-axial classification when appropriate, partial classification where semantic or conceptual ambiguity exists, and hyper specification when necessary.

These case studies validate Wikidata’s design philosophy: classification is modular, pluralistic, optional, and context-sensitive. A multi-axial system enables this by avoiding hard constraints on how many or which axes must be used for a given entity.

\section{Formal Modeling of Multi-Axiality}

Wikidata’s ontology structure can be modeled formally using tools from graph theory, which helps characterize its core architectural features: polyhierarchy, orthogonality of classification axes, and modular extensibility.

\subsection*{Graph-Theoretic View: Directed Acyclic Graphs (DAGs)}
Wikidata's subclass hierarchy (\texttt{P279}) ideally forms a directed acyclic graph (DAG).\footnote{There are actually a small number of cycles, but these cycles are rightfully treated as errors.} In contrast to tree-structured ontologies, DAGs allow nodes to have multiple parents, which results in polyhierarchies. This design permits classes to be simultaneously subclass of multiple top-level classes.

From a graph-theoretic perspective, each classification axis can be viewed as an independent subtree within the DAG. The global ontology thus consists of multiple, partially overlapping subgraphs rooted in \texttt{entity} (Q35120). This enables classification without requiring strict path uniqueness.

\subsection*{Lattice Visualisation}
When ontologies are structured along multiple orthogonal classification axes, the resulting conceptual space can be visualized as a lattice. Each axis adds a new dimension, and the number of possible class combinations grows multiplicatively as \( a \times b \times \dots \times n \), where each variable represents the number of disjoint branches in an axis. For instance, Sowa’s top-level categorization yields 12 distinct classes through combinations of his high-level splits \cite{sowa2001TopLevel}. 

In such systems, lattice visualizations are genuinely useful: they help reveal the interplay of orthogonal distinctions, expose missing combinations, and support reasoning over conceptual intersections. However, this only works when the axes involved are semantically meaningful, orthogonal, and broadly applicable. Without these conditions, the lattice becomes inflated and analytically useless.

A core assumption of our approach to Wikidata is that the introduction of a new axis, such as a disjoint \emph{apple}/\emph{non-apple} split under \texttt{entity}, should not disrupt or invalidate anything elsewhere in the ontology. And indeed, it doesn't: Wikidata’s architecture is modular enough to absorb arbitrary splits without formal contradiction. But this architectural leniency has consequences when it comes to lattice construction. Even if a newly added axis is benign in isolation, including it in a lattice can unnecessarily double or triple the number of nodes, cluttering the space with semantically trivial combinations.

This problem is especially pronounced in Wikidata, where some axes are structurally placed high in the hierarchy simply because there is no better alternative. The \emph{observable}/\emph{unobservable} distinction, for example, appears directly under \texttt{entity}, but its modeling relevance is limited and its coverage sparse. Including it in a lattice alongside more foundational splits like \emph{abstract}/\emph{concrete} or \emph{individual}/\emph{collective} leads to a dramatic increase in complexity with little ontological payoff.

For this reason, we argue that lattice visualizations in Wikidata should be selective. They should include only those axes that are philosophically motivated, widely used, and relevant across multiple domains. Otherwise, the visualization becomes not a map of meaningful intersections, but a bloated product of arbitrary modeling artifacts.

\subsection*{Orthogonality and Redundancy}

Orthogonality of classification axes implies that each axis is logically independent of the others. For example, the classification of an entity as \emph{individual} or \emph{collective} does not constrain whether it is \emph{abstract} or \emph{concrete}.

Using information-theoretic measures (e.g., mutual information), one could assess the degree of independence between axes. A high degree of correlation would suggest redundancy or conceptual entanglement, whereas low mutual information supports the notion of true orthogonality.

In practice, some of these primary splits are not truly primary, but still make sense to model as primary. For example, ``class'' is a proper subclass of ``abstract entity''. Therefore a primary split of \emph{class}/\emph{individual} is actually not primary, and the split could have been handled elsewhere. But since many ontologies treat this distinction as primary, it still makes sense to encode this classification into Wikidata.

\section{Interoperability with External Ontologies}
Wikidata is in practice a central hub of different ontologies and its classification system is designed to operate within a broader semantic ecosystem. It connects to external ontologies, vocabularies, and databases through dedicated properties \cite{vrandecic2014wikidata}.

Since Wikidata does not enforce a single upper ontology, it can integrate multiple external systems, even when their classifications conflict. In this sense, it functions as a hyperontology: a modular and semantically connected structure that supports partial alignment across diverse ontological frameworks \cite{Kutz2010CarnapGoguen}. It tolerates logical heterogeneity and treats links between ontologies as meaningful components of the system.

The subclass hierarchy (using \texttt{subclass of} (P279)) provides insufficient but still viable formal structure to support reasoning, while the external identity links offer anchors into more formally constrained ontologies. While Wikidata adopts and integrates objects from ontologies, it does not necessarily adopt all of the relationships that were typed in the ontology. This leads to differences in the ontology of Wikidata with other ontologies, making it unique.

For the upper ontology, this uniqueness can play out in various ways. Many different ontologies have equivalent classes, with differing nuances. The Wikidata community has to do the work of making sure that these classes don't create any contradictions in logical inferencing.

\subsection*{External Identifiers as Ontological Anchors}

Wikidata contains thousands of dedicated external identifier properties (e.g., \texttt{P496} for ORCID, \texttt{P699} for Disease Ontology, \texttt{P486} for MeSH) which tie items to authoritative sources.\cite{vrandecic2014wikidata} Well known items are likely to have tens of identifiers.

These identifiers are used to clarify, in cases of ambiguity, what an item means specifically. These can also be used as a bridge to connect ontologies with each other.

\subsection*{Mapping Foundational Classes}
There are some foundational classes that are not ontology-specific, but that exist in many ontologies at once. \texttt{Entity} refers to the same thing as \texttt{Thing} or \texttt{Object}. \texttt{Class} and \texttt{Type} can be used interchangeably. Even more specific classes like \texttt{Continuant} are not present in only one ontology, but many.

This means that the information on Wikidata has the potential to be reflective of external ontologies. This requires extensive work to be done on the upper ontology of Wikidata.

\section{Evaluating Axis Coverage}

\emph{Axis coverage} refers to the extent to which eligible classes in an ontology are explicitly typed along defined classification axes. A class is considered \emph{eligible} for an axis if it is a subclass of either one of the disjoint branches of the root node. For example, \texttt{painting} would not be eligible for the \emph{named}/\emph{unnamed} axis, since it is not a subclass of either class.

Ideally, all eligible classes under a given root should be typed along each relevant axis. In practice, many remain untyped or only partially classified across these dimensions. In this section, we focus on class-level typing (i.e., \texttt{P279} relations), and examine to what extent Wikidata’s classes are typed multi-axially.

All queries were run on Qlever during the last week of July 2025. 

\subsection*{Coverage under \texttt{entity}}

In Wikidata, all classes are expected to be part of the subclass hierarchy rooted at \texttt{entity} (Q35120). If a class cannot be reached from \texttt{entity} via a chain of \texttt{subclass of} (P279) statements, this is rightfully treated as an error.

Since classes are generally not explicitly marked as classes in Wikidata, a reasonable approximation is to consider any item that has at least one \texttt{subclass of} (P279) statement as a class.\footnote{This approximation includes \texttt{entity} itself, which is marked as \texttt{subclass of} “no value”, preserving its role as the root class.}

To assess ontology coverage under \texttt{entity}, we compare the total number of items that have any \texttt{subclass of} (P279) statement and the number of such items that are (recursively) subclasses of \texttt{entity} (Q35120).

\begin{verbatim}
SELECT (COUNT(DISTINCT ?item) AS ?classCount) WHERE {
  ?item wdt:P279 ?any. }
\end{verbatim}
This query reports 4239608 entities with superclasses.

\begin{verbatim}
SELECT (COUNT(DISTINCT ?item) AS ?entitySubclasses) WHERE {
  ?item wdt:P279* wd:Q35120. }
\end{verbatim}
This query reports 4210960 classes that are subclasses of \texttt{entity}, meaning that 28648 classes are not properly mapped into the Wikidata ontology. This is a moderate number and can be fixed with specific community attention.\footnote{This does not mean that there are x classes that need to be fixed individually, there are likely clumps of classes that will be fixed by connecting the top-most class to the ontology.}

\subsection*{Coverage of Abstract/Concrete Typing}
Being a primary split, and therefore exhaustive, abstract entities and concrete objects in total should also have around 4 million subclasses. There may be some classes that overlap both classes in this split, but the total number of classes like that is likely very low, especially in comparison with an axis like the \emph{named}/\emph{unnamed} axis.

These classes are disjoint, but there are some violations,\cite{DoǧanPatelSchneider2024} so our query is going to count for distinct classes. If all the violations had been resolved, the distinct query would give the same number as the non-distinct query.

To evaluate this axis, we use the following SPARQL query to find all classes that are subclasses of either side of the abstract/concrete distinction:

\begin{verbatim}
SELECT (COUNT(DISTINCT ?item) AS ?count) WHERE {
  { ?item wdt:P279* wd:Q8205328.  # subclass of concrete object
  } UNION {
  ?item wdt:P279* wd:Q7184903.  # subclass of abstract entity
  }}
\end{verbatim}
This returns 168728 classes classified by this axis. This means that an axis that is supposed to be exhaustive only covers around 4\% of the ontology. This lack can be explained by the under-theorisation of Wikidata as a multi-axial ontology. Classes being subclass of entity is enforced through constraints, but such an enforcement mechanism doesn't exist for axes separately. 

\subsection*{Multi-axially typed classes}
Beyond coverage under a single-axis, we also assess how many classification axes each class is typed along. While all splits given below are supposed to be exhaustive, some such as the \emph{named}/\emph{unnamed} axis may be too ambiguous for most classes, therefore the classes that are not explicitly typed into each classification are not necessarily erroneous. We use 5 of the 7 axes under entity, including the deprecated statements, because even though the split may be somehow problematic, the classes are still valid and relevant subclasses of entity. We didn't use the \emph{object}/\emph{property} axis, because concrete object is a subclass of object.

For brevity, only part of the query is given, the full query can be found on https://qlever.cs.uni-freiburg.de/wikidata/gAcaf8.

\begin{verbatim}
SELECT ?axisCount (COUNT(*) AS ?itemsWithThisManyAxes) WHERE {
  {
    SELECT ?item (COUNT(DISTINCT ?axis) AS ?axisCount) WHERE {
      {?item wdt:P279* wd:Q36809769. BIND(1 AS ?axis)} UNION 
      {?item wdt:P279* wd:Q16889133. BIND(1 AS ?axis)} UNION 
      {?item wdt:P279* wd:Q7048977. BIND(2 AS ?axis)} UNION 
      {?item wdt:P279* wd:Q4406616. BIND(2 AS ?axis)}
      ...
    }
  } GROUP BY ?item
  }
} GROUP BY ?axisCount ORDER BY ?axisCount
\end{verbatim}

This results in the following:
\begin{itemize}
    \item 2689244 classes are typed into one axis.
    \item 1349378 classes are typed into two axes.
    \item 73638 classes are typed into three axes.
    \item 43244 classes are typed into four axes.
    \item 61 classes are typed into five axes.
\end{itemize}

This means that multi-axial typing is practically used on Wikidata, but not to the extent of broad coverage. 

\section{New axes under entity}
The current set of axes under \texttt{entity} is both insufficient and poorly curated. At present, there are seven axes modeled using \texttt{disjoint union of} (P2738) statements under \texttt{entity}. Three of these are currently deprecated.

Wikidata allows for references and qualifiers on each statement. None of the seven axes include references to external ontologies, academic sources, or community discussions. These axes should be documented with appropriate references. These could include mappings to established ontologies (e.g., BFO, DOLCE, UFO), links to community discussions, or justification via modeling documentation.

Moreover, several foundational distinctions that are central to formal ontology design are entirely absent. For example, the foundational split between \emph{continuants} and \emph{occurrents}, which forms the basis of ontologies like BFO and OBO, is not represented under \texttt{entity}. This omission limits interoperability with ontologies that use these distinctions as their primary structuring principle, and makes mappings to BFO-derived systems difficult.

At the same time, axes like \emph{observable}/\emph{unobservable} entity, while not appearing as primary splits in external ontologies, can still serve important modeling roles in collaborative environments like Wikidata.

UFO's primary split is \emph{type} /\emph{individual}, but it includes three other important splits, modeled further down: \emph{endurant}/\emph{perdurant}, \emph{sortal}/\emph{non-sortal}, and \emph{rigid}/\emph{anti-rigid}. Wikidata could have these 4 axes, with references to articles defining UFO.

\section{Conclusion}
This paper has introduced the concept of a \emph{multi-axial mindset} for ontology design, based on Wikidata’s structurally pluralist classification model. Unlike foundational ontologies that impose a single primary split, Wikidata supports multiple overlapping axes under a shared root, allowing flexible and partial classification.

Our theoretical and empirical analysis shows that Wikidata accommodates polyhierarchical modeling through modular disjoint unions. Entities can participate in several axes without conflict, though current usage is inconsistent and axis coverage remains limited.

Future work should focus on expanding principled primary splits under \texttt{entity} and improving classification of upper-level classes. This would better realize the potential of Wikidata’s polyhierarchical structure and enhance its ontological coherence.

\bibliographystyle{splncs04}
\bibliography{references}

@book{arp2015building,
  title={Building Ontologies with Basic Formal Ontology},
  author={Arp, Robert and Smith, Barry and Spear, Andrew D},
  year={2015},
  publisher={MIT Press}
}

@techreport{masolo2003dolce,
  title={The WonderWeb Library of Foundational Ontologies and the DOLCE Ontology},
  author={Masolo, Claudio and Borgo, Stefano and Gangemi, Aldo and Guarino, Nicola and Oltramari, Alessandro},
  year={2003},
  institution={LOA-ISTC-CNR}
}

@inproceedings{niles2001ontology,
  title={Towards a Standard Upper Ontology},
  author={Niles, Ian and Pease, Adam},
  booktitle={Proceedings of the International Conference on Formal Ontology in Information Systems (FOIS)},
  year={2001}
}

@misc{sowa2001TopLevel,
    title = {Top-Level Categories},
    author={John Sowa},
    year = {2001}
}

@book{sowa2000knowledge,
  title={Knowledge Representation: Logical, Philosophical, and Computational Foundations},
  author={Sowa, John F},
  year={2000},
  publisher={Brooks/Cole}
}

@article{vrandecic2014wikidata,
  title={Wikidata: A Free Collaborative Knowledge Base},
  author={Vrandečić, Denny and Krötzsch, Markus},
  journal={Communications of the ACM},
  volume={57},
  number={10},
  pages={78--85},
  year={2014}
}

@article{DoǧanPatelSchneider2024,
  title     = {Disjointness Violations in Wikidata},
  author    = {Ege Atacan Doğan and Peter F. Patel‑Schneider},
  journal   = {arXiv preprint arXiv:2410.13707},
  year      = {2024},
}

@article{Kutz2010CarnapGoguen,
 title = {{C}arnap, {G}oguen, and the Hyperontologies: Logical Pluralism and 
Heterogeneous Structuring in Ontology Design},
 author = {Kutz, Oliver and Mossakowski, Till and L\"{u}cke, Dominik},
 journal = {Logica Universalis},
 volume = {4},
 number = {2},
 pages = {255--333},
 year = {2010},
 note = {Published online Nov 2010}
 }

@article{carnap1950,
  author = {Carnap, Rudolf},
  title = {Empiricism, Semantics, and Ontology},
  journal = {Revue Internationale de Philosophie},
  volume = {4},
  year = {1950},
  pages = {20--40}
}

@phdthesis{guizzardi2005ontological,
  title={Ontological Foundations for Structural Conceptual Models},
  author={Guizzardi, Giancarlo},
  school={University of Twente},
  year={2005}
}

@article{kishore2004computational,
author = {Kishore, Rajiv and Sharman, Raj and Ramesh, Ram},
year = {2004},
month = {01},
pages = {158-183},
title = {Computational Ontologies and Information Systems I: Foundations},
volume = {14},
journal = {Communications of the Association for Information Systems},
doi = {10.17705/1CAIS.01408}
}

@techreport{Masolo2003WonderWeb,
 title = {WonderWeb Deliverable D18: Ontology Library (final report)},
 author =
 {Masolo, Claudio and Borgo, Stefano and Gangemi, Aldo and Guarino, Nicola and 
Oltramari, Alessandro},
 institution = {ISTC-CNR, Laboratory for Applied Ontology},
 year = {2003}
 }

\end{document}